\documentclass[]{spie}  

\usepackage{amsmath,amsfonts,amssymb}
\usepackage{graphicx}
\usepackage[colorlinks=true, allcolors=blue]{hyperref}
\usepackage{subcaption}
\usepackage{colortbl}
\usepackage{booktabs}

\title{Textual Inversion for Efficient Adaptation of Open-Vocabulary Object Detectors Without Forgetting}

\author[1,*]{Frank Ruis}
\author[1]{Gertjan Burghouts}
\author[1]{Hugo Kuijf}
\affil[1]{TNO, Intelligent Imaging, the Hague, the Netherlands}

\affil[*]{frank.ruis@tno.nl}

\begin{document} 
\maketitle

\begin{abstract}
Recent progress in large pre-trained vision language models (VLMs) has reached state-of-the-art performance on several object detection benchmarks and boasts strong zero-shot  capabilities, but for optimal performance on specific targets some form of finetuning is still necessary. While the initial VLM weights allow for great few-shot transfer learning, this usually involves the loss of the original natural language querying and zero-shot capabilities. Inspired by the success of Textual Inversion (TI) in personalizing text-to-image diffusion models, we propose a similar formulation for open-vocabulary object detection. TI allows extending the VLM vocabulary by learning new or improving existing tokens to accurately detect novel or fine-grained objects from as little as three examples. The learned tokens are completely compatible with the original VLM weights while keeping them frozen, retaining the original model's benchmark performance, and leveraging its existing capabilities such as zero-shot domain transfer (e.g., detecting a sketch of an object after training only on real photos). The storage and gradient calculations are limited to the token embedding dimension, requiring significantly less compute than full-model fine-tuning. We evaluated whether the method matches or outperforms the baseline methods that suffer from forgetting in a wide variety of quantitative and qualitative experiments.
\end{abstract}


\section{INTRODUCTION}
\label{sec:intro}
Not too long ago, we could say ``Don't think about a pink elephant!'' \cite{kamath2021mdetr} or ``But can it do an astronaut riding a horse?'' \cite{ramesh2022hierarchical} to exemplify the generalization gap between humans and deep learning methods, but recent strides in vision-language models (VLM) have made these examples practically obsolete \cite{kamath2021mdetr, ramesh2022hierarchical}. While VLM have impressive zero-shot capabilities, they usually lack fine-grained understanding \cite{bugliarello2023measuring} and are often still outperformed by specialized models on specific tasks. Prompt engineering methods \cite{wei2022chain, chan_data_2022, mialon2023augmented} can come a long way in bridging this gap, though that might require writing a very long prompt for each specific case. Even that is not always enough, some kind of additional training is usually necessary for optimal results \cite{bugliarello2023measuring, li2021grounded}. 

The rich feature space of VLMs does facilitate learning new concepts from just a few examples \cite{li2021grounded}, though that comes with a few caveats. Re-training the entire model with an expanded dataset is prohibitively expensive, even for the parties that have the facilities to train the initial model. Additionally, the original training data may not be available anymore due to data retention privacy laws, dead links, or proprietary datasets. Finetuning on new data without any safeguards leads to catastrophic forgetting \cite{mccloskey1989catastrophic}. Methods such as training adapter layers mitigate the issue, but require routing the input samples to the correct set of adapter weights \cite{cohen2022my}. Most methods lose the model's original semantic generalization and compositionality capabilities, at least for the novel concepts.

In generative text-to-image literature, Textual Inversion (TI) \cite{gal2022textual} has become a popular approach to solve these issues. It involves a pretrained text-to-image generation model with frozen weights, and learning an embedding vector for a new pseudo-word, $S^*$, from a few example images. The pseudo-word is then treated as any other word to be used in textual queries for the generative model, allowing one to ask the model to generate, e.g., ``an oil painting of an $S^*$''. 

We apply a similar approach to object detection, however, there are some key differences between image generation and object detection. Whereas TI for generation focuses on a specific salient object or overarching style, in object detection there may be tens or more objects of various different classes in the same image, as well as significant within-class variations. We identify early language-vision fusion as well as a gradient that flows through a pretrained language model as key architectural properties required to make TI work for VLM object detection. Like textual inversion, the learning gradients flow through the full multimodal model, yet only affect the new tokens in their interaction with existing tokens. Instead of a single pseudo-word, we adaptively create a number of new tokens based on the number of embeddings required for initialization of the desired prompt sentence. We propose an initialization, training, and optimization strategy that maintains the VLM's capabilities such as semantics (e.g., real animals vs. toy animals) and domains (e.g., sketches vs. 3D model). Our approach enables finegrained detection and transfer of existing classes to novel domains from as few as three examples. To our knowledge there aren't any existing methods with comparable properties. Existing prompt learning methods such as CoCoOp \cite{zhou2022cocoop} mostly focus on: (1) classification as opposed to object detection, (2) automating human prompt engineering to find a context that optimizes general classification performance (akin to "a photo of a X") as opposed to new concepts or domains, and (3) abstract weights with no clear semantic meaning, as opposed to human-readable words that can be transferred to new semantic contexts (e.g. "a [concept]" vs. "a [concept] at night").

\paragraph{Our contributions are:} \textbf{1)} We demonstrate the power of textual inversion for efficient vocabulary expansion for object detection VLMs, enabling learning new concepts without forgetting. \textbf{2)} We show through extensive quantitative and qualitative experiments the impressive few-shot learning capabilities of textual inversion, and the ability to iteratively expand a pretrained VLM vocabulary without negatively impacting detection performance for known concepts. \textbf{3)} The capability to learn a new domain (e.g., aerial photos) from just three labeled images of only one class, improving performance on existing classes unseen in this new domain. \textbf{4)} The insight that a combination of early fusion and gradient flow through a language backbone pretrained through language modeling (e.g. BERT) are key architectural features for enabling newly learned concepts to be usable in new semantic contexts.

\section{Related work}
\label{sec:relwork}

\paragraph{Vision-language models (VLMs)} Open-vocabulary computer vision methods have rapidly advanced since CLIP \cite{Radford2021LearningTV} has shown that contrastive pretraining on a massive number of noisy image-text pairs can result in models with impressive zero-shot generalization capabilities. Methods such as ViLD \cite{gu2021open} and OWL-ViT \cite{minderer2022simple} have later made initial strides in distilling CLIP models to also perform object localization. GLIP \cite{li2021grounded} and GLIPv2 \cite{zhang2022glipv2} build a new architecture and pretraining strategy that has been optimized for object detection from scratch, at the time of their publication setting new state of the art results on COCO object detection \cite{lin2014microsoft}, both zero-shot and with finetuning. While these models perform well at detecting almost any everyday object, they struggle with fine-grained or less common objects or viewpoints \cite{bugliarello2023measuring}. However, pretrained VLMs can be rapidly adapted to mitigate these limitations with minimal training data\cite{li2021grounded,zhang2022glipv2}, through various finetuning techniques such as those discussed in the next section.

\paragraph{Prompt learning} Prompt learning methods \cite{lester2021power, llamaadapter2023, gao2023llamaadapterv2}, as opposed to manual rewriting of textual input prompts, introduce a small number of learnable parameters to adapt large pretrained (vision-)language models, usually as part of the input tokens of a transformer text or vision model. CoOp \cite{zhou2022learning}, for example, proposes to learn the context around the pretrained embeddings of a target class. For example, instead of manually trying to find a prompt that would optimally classify a class, such as "a picture of a car", they learn a set of context vectors around the target classes. CoCoOp extends CoOp with learnable tokens that are conditioned on the input image, allowing it to better generalize to unseen classes and domains. While prompt tuning methods usually apply to textual models, VPT \cite{jia2022visual} shows that such contextual tokens can similarly improve a vision-only model. These methods usually build on top of CLIP \cite{Radford2021LearningTV}, which is trained by aligning global image and text features, and requires massive amounts of data to adapt to localization tasks \cite{gu2021open, minderer2022simple}. GLIP \cite{li2021grounded} does perform prompt tuning \cite{lester2021power} for few-shot object detection, but their approach does not generalize to new contexts, and has the same limitations as a specialized model, in that after finetuning the model is limited to the subset of classes that it was finetuned on.

\paragraph{Mitigating forgetting}
Catastrophic forgetting \cite{mccloskey1989catastrophic} is the tendency of neural networks to forget earlier tasks when learning new tasks. Continual learning methods \cite{chen2018lifelong} aim to mitigate this by learning new tasks incrementally with minimal loss of performance, e.g. through replaying samples seen in earlier training tasks \cite{ding2022don}, or selectively slowing down learning on specific weights that are important for earlier tasks \cite{kirkpatrick2017overcoming}. Other methods aim to localize and edit sparse connections with minimal side effects, Meng et al.\cite{meng2022locating}, e.g., show one can robustly locate factual associations in a large language models and edit them by making small rank-one changes in a single MLP module, while maintaining both specificity and generalization on earlier tasks. These methods often require access to previous tasks' training data, only mitigate the problem in part, or require complex training or search strategies. We instead opt to explicitly rule out the possibility of forgetting by keeping all original model weights intact.

\paragraph{Inversion}
Inversion is a popular technique in GAN \cite{goodfellow2020generative} inversion literature \cite{zhu2016generative, zhu2020domain, isola2017image}, aiming to extract salient semantic information from images and mapping this to latent codes. Textual Inversion \cite{gal2022textual}, building on insights from PALAVRA \cite{cohen2022my}, \emph{expands} the vocabulary of a pretrained text-to-image diffusion model to be able to generate new concepts by learning from just a few examples. Similar to CLIP-based vision-language architectures, these generative methods usually focus on specific salient objects or overarching styles as opposed to many fine-grained and localized objects, and coherent visual reconstruction instead of recognition.

We propose a formulation of textual inversion adapted to object detection. While our proposed approach is generally applicable to any architecture leveraging a language model with word embeddings, \cite{li2022lavis, ShilongLiu2023GroundingDM, xu2022odise} in this work we consider the GLIP \cite{li2021grounded} model for our experiments. GLIP is, at the time of writing, the best performing open-source VLM with localization capabilities as well as early vision-language fusion. Its architecture has been built from scratch with object detection in mind, and was pretrained on millions of image and text pairs, resulting in a rich semantic feature space.

\begin{figure*}[tbh]
\centering
    \includegraphics[width=0.95\textwidth]{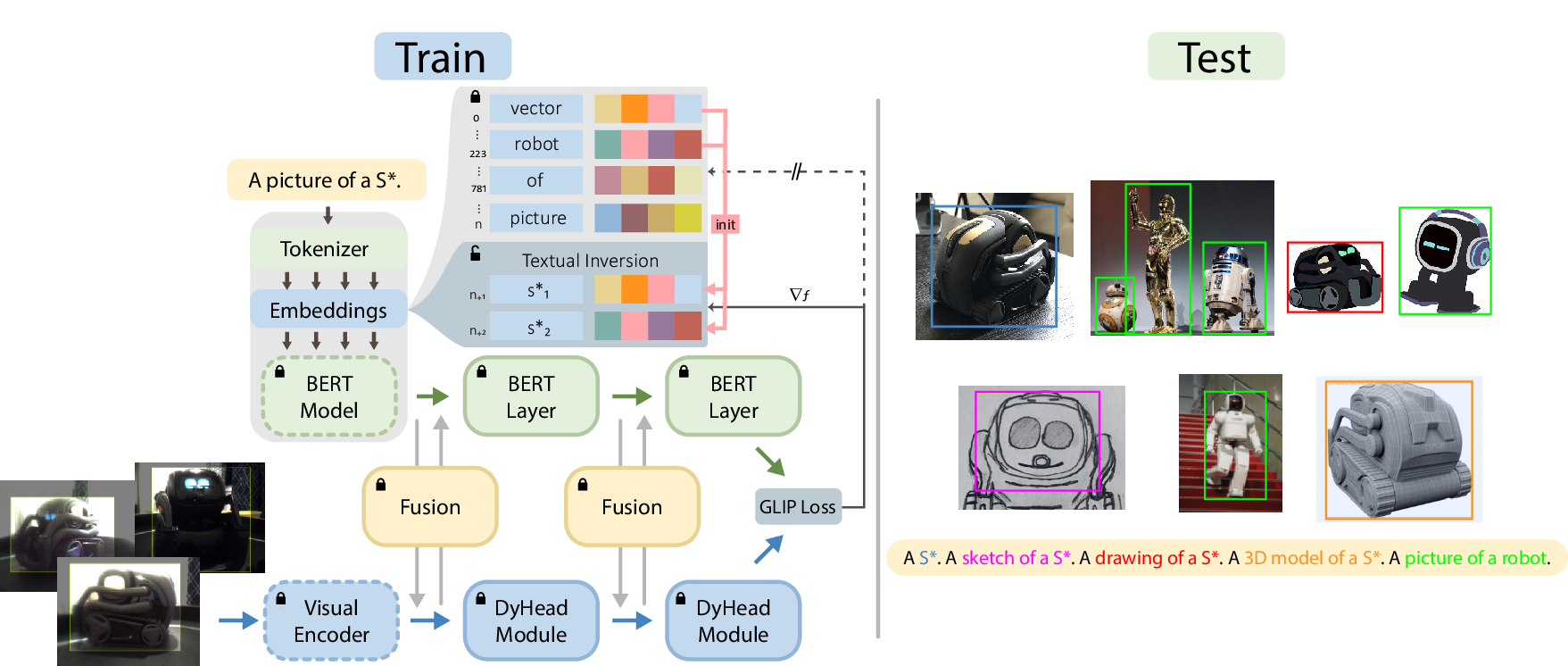}
    \caption{Textual inversion for vision-language model (VLM) object detection (here pretrained GLIP \cite{li2021grounded}). The embeddings of new tokens $s^*_n \in S^*$ are optimized during training (left) to detect a new concept, via gradients that flow through the frozen weights of the full multimodal model. Newly learned concepts ($S^*$) can be used in new contexts (right) by leveraging the base model's rich semantic knowledge (all other tokens in the prompt).}
    \label{fig:summary}
    \vspace{-1em}
\end{figure*}

\section{Method}
\label{sec:method}
In this section we show how we apply textual inversion to object detection, learning to expand the original VLM vocabulary from as few as 3 examples. The expanded vocabulary allows the model to (1) detect novel or fine-grained concepts, (2) improve the performance on existing concepts, (3) transfer to novel domains from as few as one class in the novel domain (e.g., learning to detect people from aerial viewpoints, using 3 labeled images of only cars from aerial viewpoints). Our model maintains compatibility with the original vocabulary (e.g., learned classes), and capabilities learned during pretraining (e.g., domain transfer from real photographs to sketches). 

\subsection{GLIP, prompt tuning, and architectural considerations}
\label{sec:glip}
As motivated in Section \ref{sec:relwork}, we employ GLIP \cite{li2021grounded} as our base VLM. GLIP uses BERT \cite{devlin2018bert} as its language encoder, and performs early fusion with the visual representations from a DyHead \cite{dai2021dynamic} model, allowing the language input to guide the selection of salient visual features for the object detection task and the visual information to ground the language model. We identify this last property as a key architectural feature to make textual inversion generalize to new contexts, through early experiments on other open-vocabulary detection or segmentation architectures as well as results from related work. CoCoOp \cite{zhou2022cocoop}, for example, has shown that prompt learning for classification with CLIP is not generalizable to unseen contexts unless conditioned on the input image. Since CLIP models lack such early fusion, PALAVRA \cite{cohen2022my}, e.g., needs to train a set encoder with a cycle-consistency loss on large amounts of data to be able to learn generalizable new concepts for CLIP's vocabulary. This is further evidenced by our early experiments on attempting to apply textual inversion to ODISE \cite{xu2022odise}, an open-vocabulary segmentation model built on pretrained text-to-image latent diffusion models. Although textual inversion was originally proposed for such an architecture, we were unable to learn new concepts that could generalize outside of the domain of the training data. A key difference between text-to-image diffusion models and ODISE, however, is that ODISE does not condition the image features in the U-net backbone on the textual input prompt, but on the output of an implicit captioning model instead.

The original GLIP paper proposes a version of Prompt Tuning\cite{lester2021power} as a way to fine-tune the model to specific datasets. The authors show its strong performance on 13 diverse object detection datasets with as little as 3 examples per class. The method works by learning zero-initialized delta weights for each of the input tokens, which are added to the language model output sequence just after the BERT model in Figure \ref{fig:summary}. A limitation of this learning procedure is that the gradients do not flow through the language backbone itself. Effectively, this reduces the degree of learning the semantic interactions of novel tokens with existing tokens. 

Prompt tuning is originally implemented as $P_{pt} = P + P'$, where $P \in \mathbb{R}^{M\times d}$ is the output of the language encoder $\text{Enc}_{L}(\text{Prompt})$ and $P'$ are zero-initialized weights. All weights except for $P'$ are kept frozen, so prompt tuning essentially learns a set of weight deltas on top of the input prompt to adapt to new datasets, starting at the same point as the original weights because of the zero-initialization. The encoded prompt $P$ consists of the entire input prompt, including special tokens such as the padding token. As such, the prompt can no longer be adapted, because the learned weights encompass the entire context size and are tied to specific input tokens and positions. For our experiments where we change the prompt at test time, we relax this to only learn weights for specific target tokens.


\subsection{Textual inversion for object detection}

Most existing prompt learning methods (cf. Section \ref{sec:relwork}), usually applied to CLIP, learn a set of prefix weights without semantic meaning that improve performance on a specific fine-tuning dataset. We instead aim to learn novel, semantically meaningful and generalizable, tokens that can operate in new prompts or contexts. The objective is to learn or maintain proper semantic interactions with the other input tokens in relation to the visual features. For instance: how the words `3D model' would modify the visual appearance of a target class `Anki Vector robot'. This is our motivation for placing the new tokens as early in the architecture as possible, allowing gradients to pass through the full vision-language backbone of the object detection VLM, i.e., before the BERT model in Figure \ref{fig:summary}.

TI was proposed for personalizing text-to-image generation. The text encoders as typically used in vision-language models work by first encoding a textual input sequence into a set of tokens, which is an index in a pre-defined vocabulary $V$. Each token is represented by an embedding vector $w \in W$, which is learned to reflect the semantic meaning of the text it represents. The idea of TI is simple: inject a new token $\{s^*\} \cup V$ into the language model's vocabulary, and freeze all VLM parameters except for the new token's embedding vector $w_{s^*}$. Then, train the model to recognize a new concept by showing it a small number of examples of the new concept (roughly 3-5), described by a sentence containing the new token. The learning objective is the same as that of the original model. The new token's embedding can optionally be initialized with the embedding of an existing token, e.g. starting out with the embedding for `dog' when teaching the model to recognize a specific type of dog breed.

This initialization is key for generalization to new contexts and optimal training speed and stability. Therefore, we implement an adaptive token count per new concept, based on the token count of the desired initialization. Given a vocabulary $V$ of $M = |V|$ (sub)-word tokens, encoded as learnable word embeddings $W \in \mathbb{R}^{M\times d}$, we want to learn a new concept $S^*$. Generally, if we want to initialize $S^*$ with a text that is tokenized into $n$ tokens $t_i \in V$, we would add the new tokens $\{s^*_{M+1}$, ..., $s^*_{M+n}\} \cup V$ initialized with respective word embeddings $w_{t_1}, .., w_{t_n} \in W$. For example, the novel object class `Anki Vector robot' would be assigned four novel tokens, i.e., (`$s^*_{an}$', `$s^*_{ki}$', `$s^*_{vector}$', '$s^*_{robot}$'). Since we start with relevant initial embeddings (instead of random or zero initialization), we aim to stay close to these approximately correct embeddings. For that purpose, we keep an exponential moving average of the optimized weights, which is common practice for stabilizing noisy learning signals such as those from few-shot learning with mini-batch gradient descent. See supplementary information A.2 for specific hyperparameter choices and other details.

\section{Experiments}

With our experiments we want to evaluate two main points. First, we want to show that textual inversion for object detection performs at least as well as a strong baseline (prompt tuning). Second, we want to show that properties from textual inversion in generative literature also apply to object detection. These encompass semantic compatibility with pretrained concepts, learning novel concepts, and learning from as few as 3 examples, without forgetting pretrained capabilities. All experiments were performed on an NVIDIA 3090 video card.

\subsection{Object detection in the wild}

GLIP \cite{li2021grounded} is evaluated the object detection in the wild benchmark (ODinW) \cite{li2021grounded} to assess the model's zero and few-shot learning capabilities, including prompt tuning. To show textual inversion (TI) can perform at least as well as prompt tuning, we too perform experiments on ODinW. We focus on the 3-shot splits, since that is the split used in the few-shot detection track for the extended ODinW benchmark introduced by ELEVATER \cite{li2022elevater}. It is also the point where the GLIP paper started showing diminishing returns w.r.t. adding more labeled samples. Additionally, we have repeated experiments on ODinW for prompt tuning with GLIP, since GLIP's prompt tuning weights are not available for download.

\paragraph{Results and discussion} Table \ref{tab:odinw} shows the results of our baseline experiments on ODinW. The metric is AP@0.5:0.95 unless stated otherwise. Textual inversion (TI) performs nearly identical to prompt tuning (PT) on average. The performance on individual datasets is comparable as well, with a couple of differences. We report the TI results with EMA disabled for some of these differences (Table \ref{tab:odinw}, bottom row), showing that those results are much more similar to prompt tuning, both in the positive direction (packages dataset) and in the negative direction (shellfish dataset). 

\begin{table}[tb]
\caption{Zero shot, full tuning, and 3-shot object detection on the ODinW benchmark. The first three rows are taken from the GLIP paper \cite{li2021grounded}, while $\dagger$ is our reproduction. }
\label{tab:odinw}

\begin{tabular}{@{}llllllll}
                           & \multicolumn{1}{l}{\textbf{VOC}} & \textbf{Aerial} & \multicolumn{1}{l}{\textbf{Aquarium}} & \textbf{Rabbits} & \textbf{EgoHands} & \textbf{Mushroom} & \textbf{Packages} \\ 
                           \toprule
\textbf{GLIP-L (zero)}  & 61.7 & 7.1 & 26.9 & 75.0 & 45.5 & 49.0 & 62.8 \\
\textbf{GLIP-L (full$^*$)} & 69.6 & 32.6 & 56.6 & 76.4 & 79.4  & 88.1 & 67.1  \\
\textbf{GLIP-L PT}              & $65.0_{\pm 0.5}$                             & $21.4_{\pm 1.0}$          & $43.6_{\pm 1.1}$                            & $72.9_{\pm 0.7}$       & $70.4_{\pm 0.1}$    & $91.4_{\pm 0.7}$   & $57.7_{\pm 3.7}$  \\
\textbf{GLIP-L PT$^\dagger$}              & $66.1_{\pm 2.5}$                             & $22.3_{\pm 2.0}$          & $46.8_{\pm 1.0}$                            & $72.5_{\pm 1.3}$       & $70.0_{\pm 1.0}$        & $83.1_{\pm 2.0}$  & $72.5_{\pm 1.3}$     \\ \midrule
\textbf{GLIP-L TI} & $68.5_{\pm 0.4}$          & $21.2_{\pm 1.8}$                & $47.8_{\pm 2.2}$       & $74.5_{\pm 0.3}$         & $68.3_{\pm 0.4}$           & $83.1_{\pm 2.1}$  & $66.6_{\pm 0.3}$     \\
- no EMA & -          & -                & -       & $72.7_{\pm 1.2}$         & $70.1_{\pm 0.2}$           & -  & $73.4_{\pm 1.0}$     \\ 
\\
                           & \textbf{Raccoon} & \textbf{Shellfish} & \textbf{Vehicles} & \textbf{Pistols} & \textbf{Potholes} & \textbf{Thermal} & \textbf{Average} \\  \toprule
\textbf{GLIP-L (zero)}  & 63.3 & 68.9 & 57.3 & 68.6 & 25.7 & 66.0 & 52.1 \\
\textbf{GLIP-L (full$^*$)} & 69.4 & 65.8 & 71.6 & 75.7 & 60.3 & 83.1 & 68.9 \\
\textbf{GLIP-L PT}   & $70.7_{\pm 1.1}$        & $69.7_{\pm 0.9}$        & $62.6_{\pm 0.8}$        & $67.7_{\pm 0.4}$       & $36.2_{\pm 1.1}$        & $68.8_{\pm 1.5}$       & $61.4_{\pm 0.3}$ \\
\textbf{GLIP-L PT$^\dagger$}  & $62.3_{\pm 4.0}$        & $63.0_{\pm 4.0}$        & $66.8_{\pm 1.0}$        & $67.2_{\pm 2.2}$       & $34.0_{\pm 3.0}$        & $77.8_{\pm 0.5}$       & $61.9_{\pm 1.2}$ \\ \midrule
\textbf{GLIP-L TI}  & $62.1_{\pm 3.1}$     & $67.7_{\pm 1.3}$       &  $67.4_{\pm 0.9}$      & $68.6_{\pm 1.7}$             & $36.8_{\pm 3.3}$   & $76.1_{\pm 2.7}$             & $62.2_{\pm 1.6}$ \\
- no EMA & -          & $63.5_{\pm 3.5}$        & -       & -         & -           & -  & -     \\ \bottomrule

\end{tabular}
\footnotesize{* full tuning tunes all model weights on the entire dataset.}
\end{table}

From inspecting the training data (cf. supplementary figure 2), we find that these discrepancies are mainly caused by significant label noise. In some cases it is more optimal to stay close to the initial weights (EMA), while in others it is more beneficial to be able to rapidly learn from the easy correctly labeled samples before the noisy samples start negatively affecting the results (no EMA). One of the shellfish dataset splits, e.g., has more crabs labeled as lobsters than correctly labeled lobsters or crabs, and one of the crab labels is a picture of a crab cake while the rest of the dataset does contain actual crabs. As there are only 3 images per class, this explains why neither method performs better than zero-shot, as well as the large variance over the different splits. In general, both methods come surprisingly close to finetuning the full model on the entire training set. We refer the reader to the GLIP \cite{li2021grounded} and GLIPv2 \cite{zhang2022glipv2} papers for an even more comprehensive evaluation of prompt tuning.

Note that our reproduction$^\dagger$ of the prompt tuning results leads to slightly higher performance on average than was reported in the GLIP paper, but there are some significant deviations in both directions for the individual datasets. We suspect this may be due to the inherent noise of few-shot learning on noisy datasets mentioned above, as well as a possible difference in the dataset or optimization parameters that is not reflected in the ODinW configuration files. For example, the `Chicken of the Woods' mushroom in the `Mushrooms' dataset has its class name shortened to `CoW' in the configuration files, which would be initialized with the embedding of the mammal instead of the mushroom.

\subsection{Effect of new tokens on existing tokens}

TI explicitly does not affect the pretrained model weights, but newly learned tokens do interact with existing tokens when both are included in the same prompt. During pretraining there is an abundance of data and each token is used in many different contexts, but when learning TI tokens from a few examples this is not the case. When training on the three examples in Figure \ref{fig:summary} for example, a shortcut that learns to detect the color dark gray could still reach perfect training accuracy, but would introduce large amounts of false positives in other contexts. Therefore, we investigate whether the TI tokens, which work well for the new classes, may degrade performance when used together with known tokens from the base model. 

We assess the TI tokens against all other classes of all other ODinW datasets, by including the other classes in the prompt as distractors. The aim is to have a low degradation in case of such distractors. That would imply that forgetting is not severe after adding new tokens. We look at two sets of weights. GLIP-L (zero-shot): The original model weights are used for the current dataset's target classes. TI: 3-shot textual inversion embeddings from the previous experiment are used for the target classes. We want to compare both in terms of degradation when adding distractors. We aim for two objectives: (1) the overall performance of TI is better than GLIP-L, and, (2) the degradation for new TI tokens is not larger than GLIP-L tokens. We do this for three prompt settings. Base: Only the current dataset's classes are included in the prompt. Existing Distractors: All 43 unique classes from the other ODinW datasets are included in the prompt, using the original model's embeddings. Inversion Distractors: Same as Existing Distractors, but with the TI embeddings instead of the original embeddings.

\paragraph{Results and discussion} Table \ref{tab:distractors} shows the overall performance of GLIP-L and our TI extension, for all 13 ODinW dataset combined (top row), and the degradation when adding distractors in the prompt (second and third rows). We note that part of the degradation is due to non-exhaustive labeling (e.g., the Raccoons dataset contains unlabeled people, and `person' is one of the distractor classes). The TI tokens degrade in a similar way to the existing tokens instead of causing a collapse. This is a nice property: it means the degradation is inherent to multi-class classification and not due to TI specifically. This shows one can extend the model vocabulary incrementally, training new tokens one dataset at a time, and expect the new tokens to behave in a similar way to existing tokens learned during pretraining. For fine-grained concepts, post-hoc training may even be optimal in current VLMs, with Bugliarello et al.\cite{bugliarello2023measuring} showing that during pretraining ``performance fluctuates substantially on many fine-grained tasks, never converging''. For both zero-shot and TI the performance drop is slightly less when TI tokens are used as distractors compared to existing tokens. We suspect this is a result of the TI tokens being more specialized. Interestingly, our TI vocabulary with distractors still outperforms the zero-shot results from GLIP-L without distractors.

\begin{table}[tb]
    \caption{Robustness against distractors in the ODinW benchmark. Degradation for our TI tokens is comparable to existing GLIP-L tokens, indicating that TI tokens are as robust as those learned during large-scale pretraining. Our TI with distractors outperforms the zero-shot results without distractors.}
    \label{tab:distractors}
    \centering
    \scalebox{0.85}{
    \begin{tabular}{l|ll}
                                        & \multicolumn{2}{c}{\textbf{Average ODinW AP}} \\
                                        & GLIP-L (zero-shot)        &    TI (ours) \\ \hline
        \textbf{Base}     & 53.2   & 61.9      \\
        \textbf{Existing Distractors} & 48.5 \,\,\, \textcolor{red}{-8.8\%} &  56.3 \,\,\,\textcolor{red}{-8.9\%} \\ 
        \textbf{Inversion Distractors}       & 48.8 \,\,\, \textcolor{red}{-8.2\%} & 57.2 \,\,\,\textcolor{red}{-7.6\%} \\ \hline
    \end{tabular}
    }
\end{table}

\subsection{Fine-grained object detection} 

A common reason to train a specialist model is to perform fine-grained object detection, e.g., various different dog breeds. While the base VLM may have seen most dog breeds individually, they are likely often described by their supercategory `dog', or are rarely depicted with multiple other dog breeds. To learn fine-grained discrimination, we train textual inversion and prompt tuning on the Oxford-IIIT Pet dataset, \cite{parkhi12a} which consists of 37 classes of dog and cat breeds. The bounding boxes are quite unconventional, marking just the animal's head instead of its entire body, which adds to the challenge for zero-shot detection (cf. supplementary figure 1). The prompt used during training and inference is `<class\_1> head. ... . <class\_37> head.', where only the class tokens are learned for textual inversion and prompt tuning.  

\paragraph{Results and discussion} The AP on the Oxford-IIIT Pet dataset is:\\  

\begin{tabular}{ll|ll|ll}
    \textbf{GLIP-L (zero-shot)}     & 2.8      &  \textbf{GLIP-L PT} & $47.5 \pm 2.3$ & \textbf{GLIP-L TI (ours)}       & $47.9 \pm 1.0$
\end{tabular}\\  

Given the fine granularity and unconventional bounding boxes, it is not surprising that the zero-shot model performs close to random at this task. Prompt tuning and TI, however, are able to reach $\sim$48\% AP with just 3 examples per class. We evaluated various TI token initialization strategies, including adaptive tokens, the super-category (dog or cat), random init, and random words (e.g., potato). We report the adaptive tokens here, as all results were within 2\% variation. The main benefit of adaptive token initialization is in the ease of optimization and convergence speed: To reach the same performance as adaptive initialization, all non-adaptive initialization strategies required at least three times as many training epochs, and several learning rate drops after validation performance had plateaued. However, in the next section we show that the initialization strategy is important for performance when using the newly learned tokens in new semantic contexts.

\subsection{New tokens combined with prior semantic knowledge} 

One objective of our method is to avoid forgetting of learned capabilities, including semantics, such as real animals vs. stuffed animal toys. To assess this capability, we improve the model's concept of a corgi dog by training on the corgi images from the DreamBooth \cite{ruiz2022dreambooth} dataset (located in the `dog6' folder\footnote{\url{https://github.com/google/dreambooth/tree/main/dataset/dog6/}}). We take 3 images for training and use the remaining two for validation. We train textual inversion and prompt tuning with the prompt `A picture of a corgi puppy', where `corgi puppy' are the tokens to be adapted. We test on `A stuffed animal of a corgi'. 

\begin{figure*}[tb]
\centering
    \includegraphics[width=0.75\textwidth]{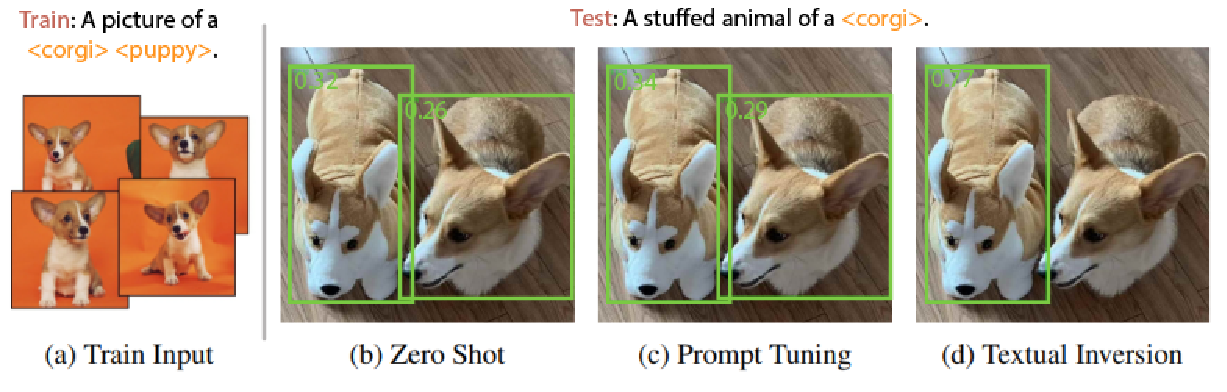}
    \caption{Qualitative example of domain transfer. At test time we attempt to detect just the stuffed animal on the left, not the real animal on the right. Only textual inversion is successful at this task.}
    \label{fig:corgiqual}
\end{figure*}

\paragraph{Results and discussion} Figure \ref{fig:corgiqual} shows an example of predictions for an image containing both an adult corgi and a stuffed animal of a corgi, where we attempt to detect just the stuffed animal. GLIP-L zero-shot and prompt tuning are not able to disambiguate when both are present in the same image. TI, on the other hand, seems to be able to leverage its improved concept of a corgi to now confidently detect just the stuffed animal, a concept which it had already learned during pretraining.

\begin{figure*}[tb]
\centering
    \includegraphics[width=0.75\textwidth]{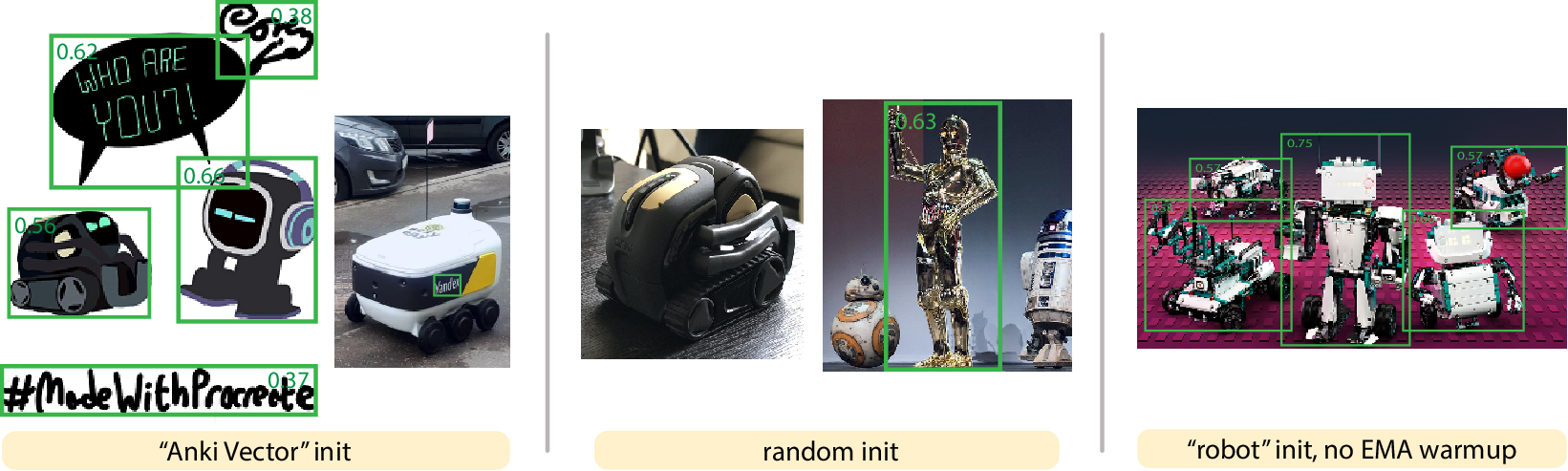}
    \caption{Typical examples of detection mistakes encountered for different choices of initialization and training strategies. Each example shows model predictions for the textual prompt "An $S^*$", where $S^*$ erroneously detects vector graphics (left), misses the target class and erroneously detects a different robot (middle) or detects any robot instead of only the target class (right). }
    \label{fig:botinits}
\end{figure*}

\subsection{New tokens combined with domain transfer} 

We also aim to maintain the pretrained base model knowledge about domains, e.g., sketches vs. 3D models vs. real photos. For that purpose, we learn tokens for an `Anki Vector' robot, using a small set of images from a Roboflow Universe dataset, including a couple of negative examples for `robot' initialized tokens, since they would otherwise start with perfect accuracy and have no reason to change the initial embedding. The dataset and concept were chosen for the following properties: The Anki Vector is popular enough that there is a wide variety of uniquely identifying imagery available in various domains such as drawings, 3D models, and photos. At the same time, we could not find a prompt that would robustly detect just the Anki Vector. Finally, the dataset is challenging through its low camera quality and multiple views of the robot (front, side, back). We also evaluate different token initialization options, such as `robot', `vector', and random init, as well as an EMA warmup to allow the learned tokens to more quickly deviate from the initialization token for the first few training batches.

\paragraph{Results and discussion} We find that when the new concept is not yet known by the base model, extra care needs to be taken for token initialization. For example, initializing $S^*$ with the robot's name, `Vector', detects the new concept well but retains information about vector graphics. Random initialization has difficulty generalizing to out of domain data, e.g., missing the vector sometimes while detecting non-vector objects. Initializing with `robot' but starting with EMA with high decay value will stick too close to the initial value, and keep detecting any kind of robot. We show some typical examples of these cases in Figure \ref{fig:botinits}. However, good results can be obtained with a token initialization of `robot' when warming up the decay value from 0 to 0.999 over 20 training batches, see Figure \ref{fig:summary} on the right.

\subsection{Learning a new domain from one class} 

GLIP excels at domains that are most prevalent in its pretraining data, but performance drops on out-of-domain viewpoints such as those in drone-captured imagery. In this experiment, we want to learn such a new domain (in this case `aerial') from just 3 labeled examples for only 1 class, e.g., `car'. Instead of learning the token for the target class, we learn the token for the domain (the context). We then aim to detect a known class, e.g. `person', in the newly learned domain. This would imply that we can learn a new domain from just a few images of one class, which would generalize to other classes in that new domain. Specifically, this setup evaluates if the method can (a) improve detection performance without leaking information about the `car' class into the `aerial' token, and, (b) without learning the remaining classes as hard negatives even though they might be included as background class in the three training examples. Such properties are beneficial when one wants to incrementally learn new concepts without exhaustively labeling all existing and possible future concepts.

We perform this experiment on the VisDrone \cite{visdrone2021} dataset, which is a large dataset for small object detection consisting of drone-captured images of vehicles and people. Because the dataset is fine-grained and we are not learning the target classes, we merge the person and people classes, as well as the tricycle and bicycle classes, resulting in 7 classes total. We train on three images where we discard all labels except for the `car' class, and validate on a validation set of three different images from the training set. We test on the VisDrone validation set with all other classes (since we do not use the validation set during training).

\begin{figure*}[tbh]
\centering
    \includegraphics[width=0.85\textwidth]{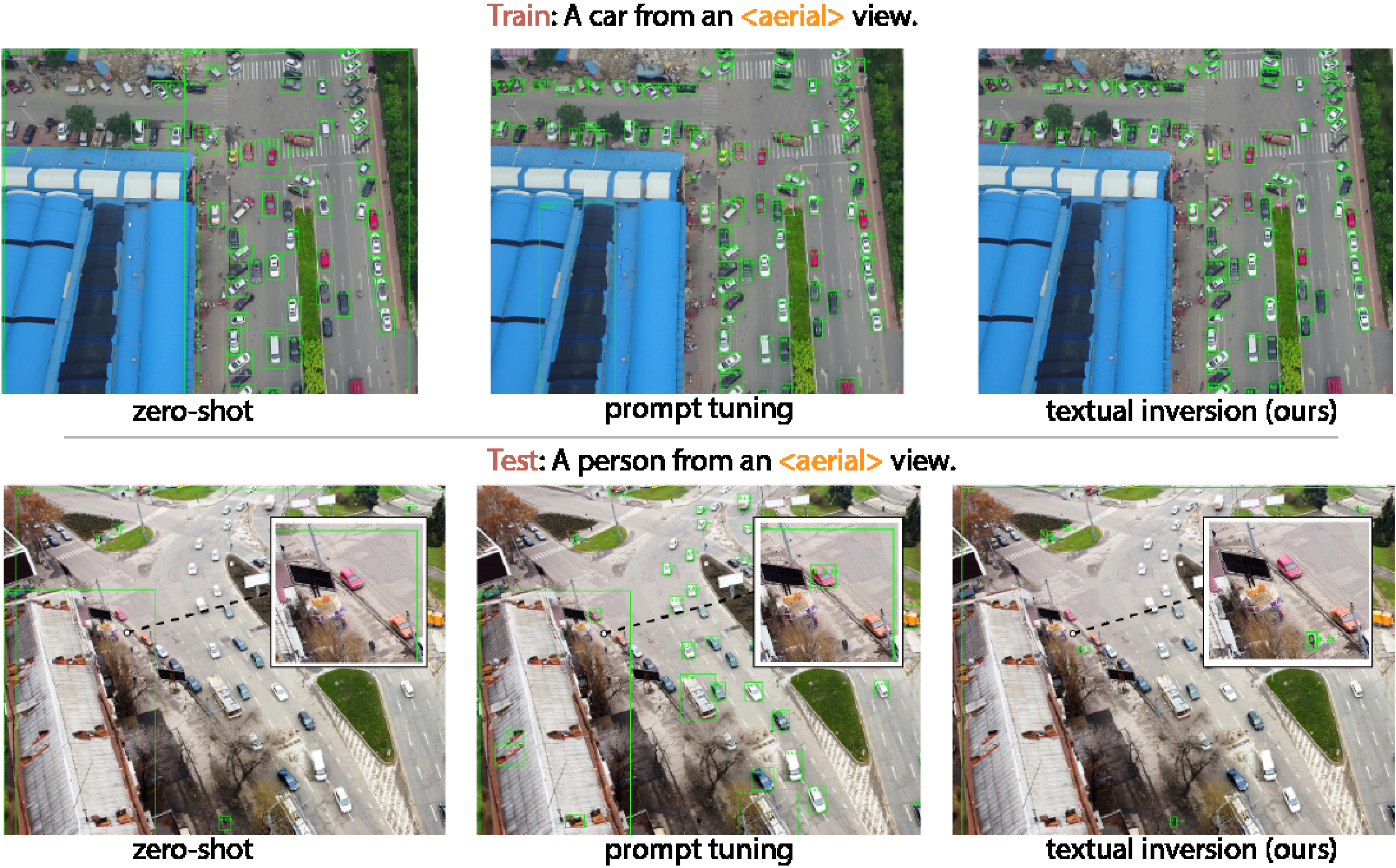}
    \caption{Results for the zero-shot context transfer task on VisDrone. This is a hard task for GLIP-L and prompt tuning. Our TI vocabulary successfully learns to detect classes such as `person' and `truck' from aerial viewpoints, while only having learned the context `aerial' from 3 examples of cars.}
    \label{fig:visdrone}
\end{figure*}

\paragraph{Results and discussion} Figure \ref{fig:visdrone} shows a qualitative example of model predictions of `car' on one of the three training images (top row), as well as predictions on the unseen `person' class on an unseen image that is not part of the VisDrone dataset (bottom row). Both prompt tuning and textual inversion can improve recall and bounding box tightness on the class seen during training. However, when switching the target class from `car' to `person' during testing, prompt tuning mainly detects cars, suggesting that information about the car class has leaked into the `aerial' token. Our TI vocabulary can detect most of the people in the image with improved bounding boxes compared to the zero-shot detections.

\begin{figure*}[tbh]
\centering
    \begin{subfigure}[b]{0.42\textwidth}
        \includegraphics[width=\textwidth]{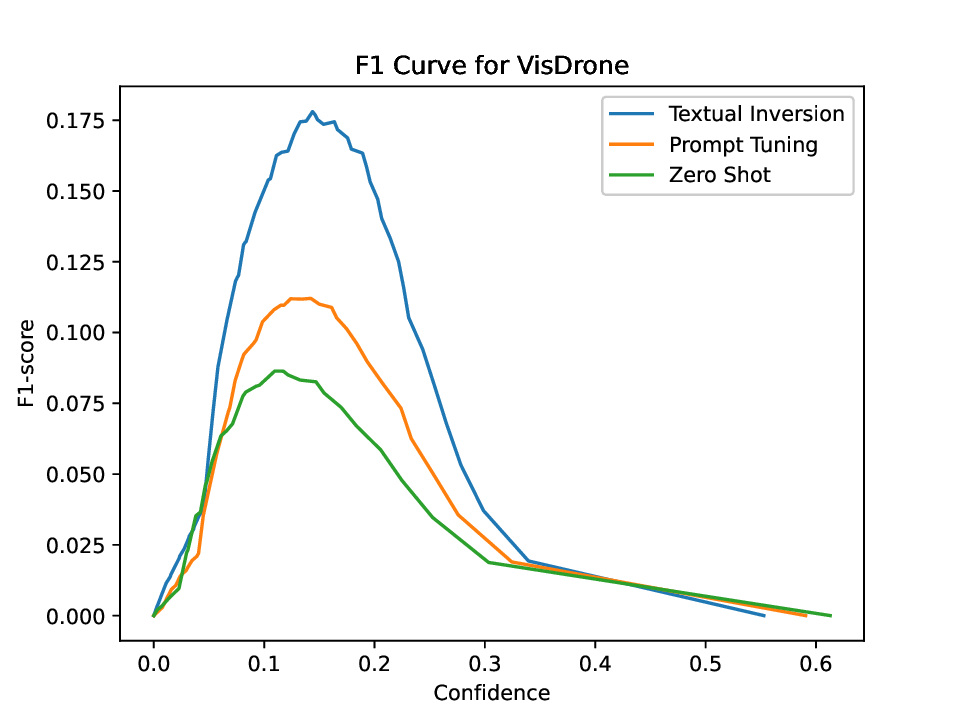}
        \caption{F1 curve}
        \label{fig:f1-cars}
    \end{subfigure}
    ~
    \begin{subfigure}[b]{0.42\textwidth}
        \includegraphics[width=\textwidth]{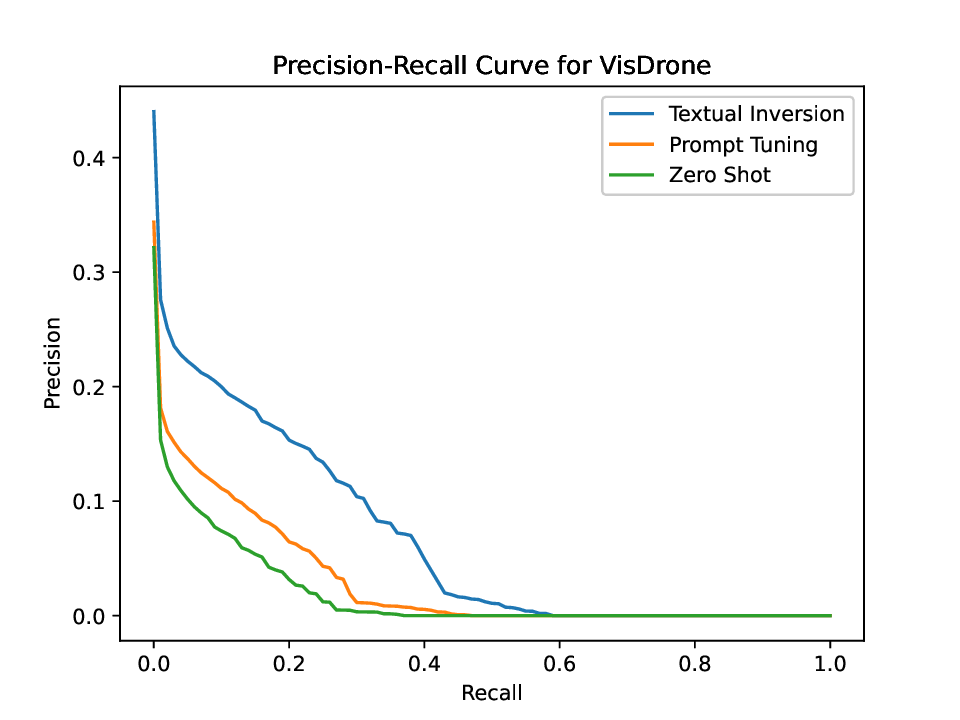}
        \caption{Precision-Recall Curve}
        \label{fig:pr-cars}
    \end{subfigure}
    \caption{Quantitative results for the zero-shot context transfer task on VisDrone.}
\end{figure*}

Figure \ref{fig:f1-cars} and \ref{fig:pr-cars} show the quantitative results for the zero-shot context transfer task, in the form of F1-confidence and precision-recall curves, for the validation set of VisDrone with the `car' class removed (since we did not use the validation set during training, and the official test set has no labels due to being part of a comptetition). The area under the curve between textual inversion and prompt tuning shows that textual inversion has a much greater freedom in the choice of confidence values that still give a relatively high recall or f1-score. This is especially important when one does not have access to a comprehensive and exhaustively labeled validation set like we used here, but need to choose a confidence value based on just a few examples. It also explains why in the qualitative example we could not find a confidence threshold for prompt tuning where people would be detected but vehicles would not. Our intuition is that textual inversion is better able to optimize the context token while staying close to its semantic meaning, due to the learnable part of the embeddings going through the language backbone. For example, the optimizer may find it more difficult to add `car' information to the `aerial' token than a different solution that makes more sense semantically in the frozen language model weights. The learnable part of prompt tuning embeddings only go through a set of BERT layers that were not pretrained through language modeling, and as such the shortcut to leak information about cars into the `aerial' token may be more optimal.

\section{Limitations}
Because textual inversion is primarily a search in an existing model's latent space, it also inherits the base model's limitations and impacts. We further elaborate on some of these limitations for our base model GLIP \cite{li2021grounded} in supplementary information A.1. The authors of GLIPv2 \cite{zhang2022glipv2}, however, show that prompt tuning becomes significantly better at few-shot learning as the base model improves. Our experiments comparing to prompt tuning suggest the same will be the case for textual inversion. Similarly, as of yet GLIP is to our knowledge the only open-source open-vocabulary object detection model that employs vision-language fusion, and as such the only available architecture that we could test our approach on.

In this work we focus on demonstrating the viability of textual inversion for expanding the vocabulary of a VLM object detector, as well as the key properties a base model should have to make this work. As such, we have considered finding the optimal augmentation and regularization settings out of scope, since those are often highly model and dataset dependent. While the base model parameters have been optimized for pretraining on millions of images and therefore need little augmentation or regularization, for few-shot learning this is the opposite, \cite{steiner2022train} so we expect much gains to be made when optimizing this for specific applications.

\section{Conclusion}
\label{sec:discussion}

We proposed Textual Inversion (TI) for object detection by integration with Vision-Language Models (VLMs). Early language-vision fusion combined with a gradient that flows through a pretrained language model are key architectural properties required to make TI work for VLM object detection. We proposed an adaptive initialization, training, and optimization strategy that maintains the VLM's capabilities such as semantics (e.g., real animals vs. toy animals) and domains (e.g., sketches vs. 3D model). We demonstrated good performance on few-shot object detection, including finegrained classes. We showed that it is even possible to transfer existing classes to novel domains from as few as three examples of just one class. Importantly, our method suffers significantly less from forgetting the rich capabilities of large pretrained VLMs.

\section{Acknowledgements}
The authors received support from the FaRADAI project (ref. 101103386) funded by the European Commission under the European Defence Fund.

\appendix    

\bibliography{report} 
\bibliographystyle{spiebib} 

\section{Supplementary Information}

\subsection{GLIP Predictions in Uncertain Feature Space}
We notice that all GLIP-based methods occasionally make high-confidence errors such as bounding boxes encompassing the entire image, or objects that do not relate to the target prompt in any way, likely stemming from the self-supervised pretraining of the base model. From our experience with GLIP, this seems to happen mainly when operating in an area of high uncertainty. In the zero-shot context transfer experiment, e.g., due to the compounding effect of few-shot learning on images that are out of domain of the pretrained model, and then performing zero-shot transfer to classes that were included as negative examples during training. These examples were purposefully chosen to be difficult for the models, and model calibration will be better when including more training data or staying closer to the domain of the training data for the base model, or using a more powerful base model such as GLIPv2.

\subsection{Hyperparameters}
For exponential moving average (EMA), we use a decay of $0.999$, sometimes with a warm-up from 0 to $0.999$ over the course of $20-500$ batches for novel concepts that the base model struggles with (detailed in the Experiments). Unless stated otherwise, we use the same learning objective and configuration parameters as proposed in the original GLIP paper for prompt tuning, where they use a learning rate of $0.05$ and a weight decay of $0.25$ at a batch size of $4$. We reduce the learning rate to $0.025$ since we can only fit a batch size of $2$ on our hardware.

Our framework of choice, PyTorch \cite{paszke2019pytorch}, does not allow individual word embeddings to require gradients while the rest do not, so we introduce an embedding manager that wraps the word embedding layer, keeps track of TI embeddings, and listens for newly added TI tokens. Figure 1 in the main paper shows how we implemented TI in the GLIP architecture.

\subsection{Training Data and Predictions}
\begin{figure*}[!hbt]
    \centering
    \includegraphics[width=\textwidth]{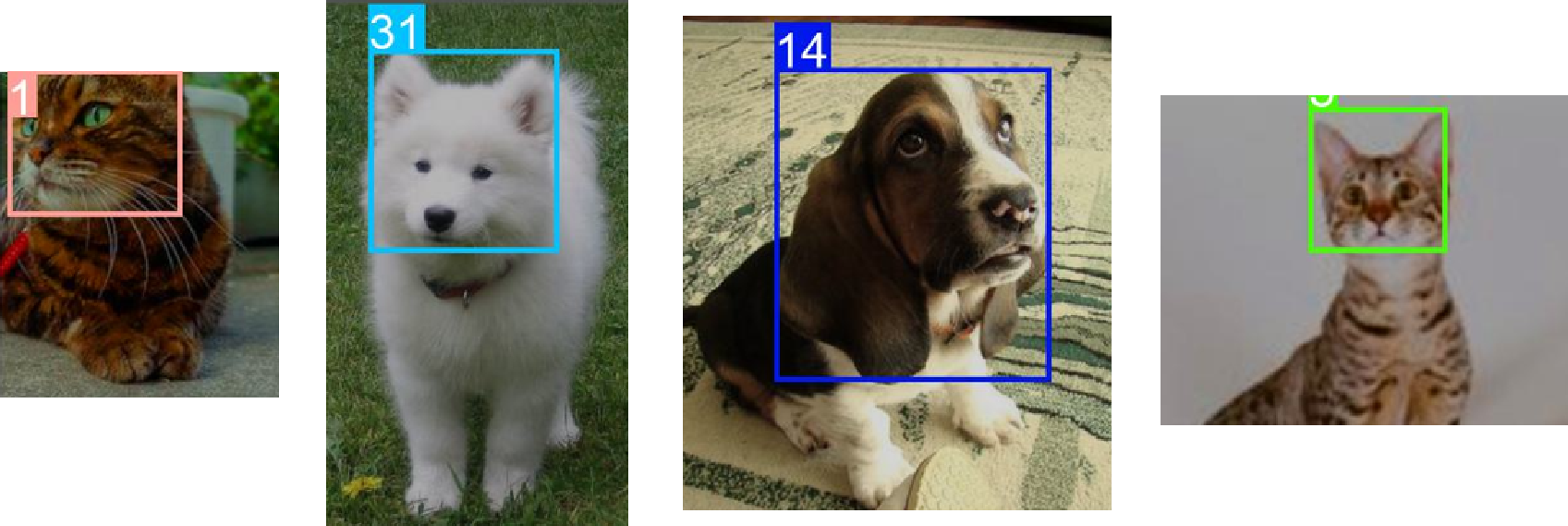}
    \caption{Some examples of bounding box targets in the Oxford Pets-IIIT dataset, which consists of 37 classes of dog and cat breeds. The bounding boxes are quite unconventional, marking just the animal's head instead of its entire body, which adds to the challenge for zero-shot detection.}
    \label{fig:petsexamples}
\end{figure*}

\begin{figure*}[!hbt]
    \centering
    \includegraphics[width=\textwidth]{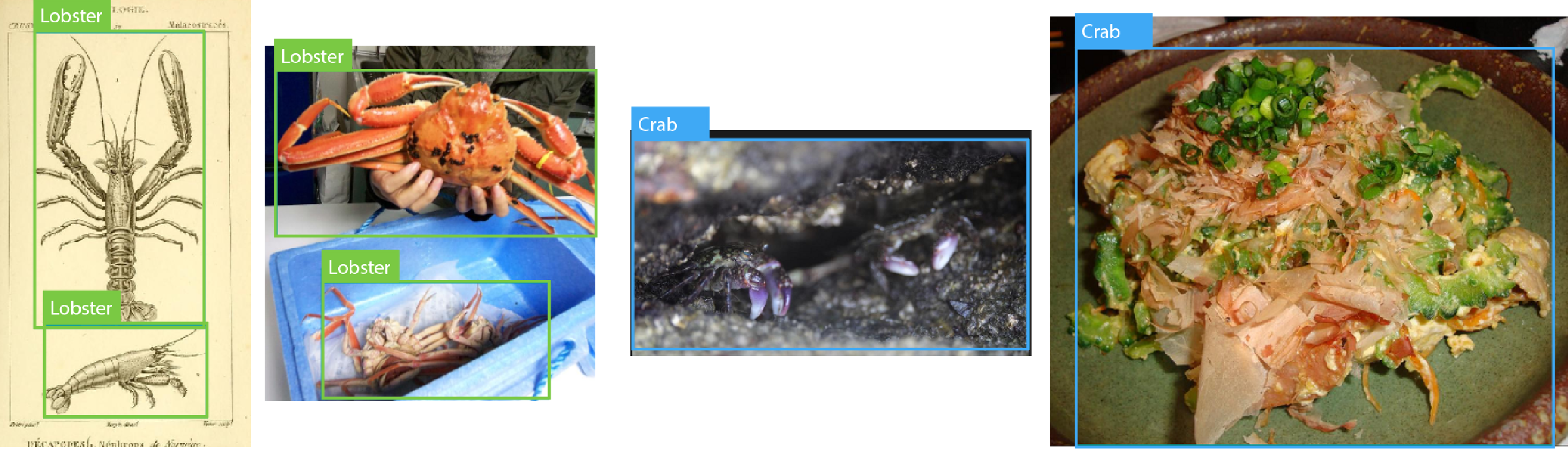}
    \caption{Some examples of label noise in the ODinW Shellfish dataset from a single split. More crabs have been labeled as lobsters than correctly labeled lobsters or crabs, and one of the crab labels is a picture of a crab cake while the rest of the dataset does contain actual crabs.}
    \label{fig:shellfishexamples}
\end{figure*}

In this section we show some representative examples of training data for the Oxford-IIIT Pets dataset (Figure \ref{fig:petsexamples}) and some examples of label errors in one of the Shellfish dataset splits from ODinW (Figure \ref{fig:shellfishexamples}).

\end{document}